%% file: main.tex
\documentclass[manuscript]{acmart}
\usepackage{xcolor}
\usepackage{wrapfig}

\AtBeginDocument{%
  \providecommand\BibTeX{{%
    \normalfont B\kern-0.5em{\scshape i\kern-0.25em b}\kern-0.8em\TeX}}}

\copyrightyear{2024} 
\acmYear{2024} 
\setcopyright{rightsretained} 
\acmConference[FAccT '24]{The 2024 ACM Conference on Fairness, Accountability, and Transparency}{June 3--6, 2024}{Rio de Janeiro, Brazil}
\acmBooktitle{The 2024 ACM Conference on Fairness, Accountability, and Transparency (FAccT '24), June 3--6, 2024, Rio de Janeiro, Brazil}\acmDOI{10.1145/3630106.3658999}
\acmISBN{979-8-4007-0450-5/24/06}


\begin{document}

\title{SIDEs: Separating Idealization from Deceptive `Explanations' in xAI}


\author{Emily Sullivan}
\email{e.e.sullivan@uu.nl}
\orcid{0000-0002-2073-5384}
\affiliation{%
  \institution{Utrecht University}
  \city{Utrecht}
  \country{Netherlands}
}




\begin{abstract}
{Explainable AI (xAI) methods are important for establishing trust in using black-box models. However, recent criticism has mounted against current xAI methods that they disagree, are necessarily false, and can be manipulated, which has started to undermine the deployment of black-box models. Rudin (2019) goes so far as to say that we should stop using black-box models altogether in high-stakes cases because xAI explanations ‘must be wrong’. However, strict fidelity to the truth is historically not a desideratum in science. Idealizations--the intentional distortions introduced to scientific theories and models--are commonplace in the natural sciences and are seen as a successful scientific tool. Thus, it is not falsehood \textit{qua} falsehood that is the issue. In this paper, I outline the need for xAI research to engage in \textit{idealization evaluation}. Drawing on the use of idealizations in the natural sciences and philosophy of science, I introduce a novel framework for evaluating whether xAI methods engage in successful idealizations or deceptive explanations (SIDEs). SIDEs evaluates whether the \textit{limitations} of xAI methods, and the distortions that they introduce, can be part of a successful idealization or are indeed deceptive distortions as critics suggest. I discuss the role that existing research can play in idealization evaluation and where innovation is necessary. Through a qualitative analysis we find that leading feature importance methods and counterfactual explanations are subject to idealization failure and suggest remedies for ameliorating idealization failure.}





\end{abstract}





\maketitle

\section{Introduction} \label{intro}
The ideal gas law dating back to 1834, along with its simpler cousin Boyle’s law from the 1660s, are still used to explain how gases behave. Boyle's law captures the inverse relationship between pressure and volume at constant temperatures, which can explain why a balloon's inflation level changes amidst elevation changes. The ideal gas law complicates the picture by adding the influence of temperature changes and molarity to gas behavior. However, both these laws involve distortions and departures from the truth \cite{potochnik2017idealization, elgin2017true}. Real gases do not behave ideally. Particles are assumed not to interact (even though they do), and the actual relationship between pressure and volume is more complicated than either law lets on. Despite this, the ideal gas law is still highly successful--a highly successful idealization (i.e. an intentional distortion introduced to scientific theories and models). 

Contrast this with a more controversial contemporary case. In 2021, RIVM, the Dutch health institute, constructed a model to measure the spread of nitrogen pollution. In order to reduce computational complexity, the model treated nitrogen deposits coming from roads as spreading only 5km, while the deposits from farms traveled much longer distances. However, this distortion--this idealization--was not a success, and was later removed because it disproportionately put the pollution blame on farms instead of highways \cite{veldhuizen22}. What makes the ideal gas law a permissible, even desirable, distortion, but the idealizations in the Dutch nitrogen model problematic, needing revision? It cannot simply be that the Dutch nitrogen model contains falsehoods \textit{vis a vis} an idealization, or a lack of fidelity to the phenomena, since the ideal gas law does the same.

Now consider the topic at hand: explainable AI (xAI) methods. \citet{rudin2019stop} argues that the increasing trend of using black-box ML models across science and society is problematic, precisely because the methods we use to interpret these models provide us with necessarily wrong explanations of how they work. Other recent works have painted a picture of mounting criticism that leading xAI techniques are unreliable, subject to manipulation, and different techniques often disagree \cite{slack2020fooling, slack2021counterfactual, dombrowski2019explanations, krishna2022disagreement, alvarez2018robustness}.
Some critics argue that xAI is a `false hope' \cite{ghassemi2021false}, where methods can engage in explanation hacking \cite{sullivan2024hacking} or fairwashing \cite{aivodji2019fairwashing}, leading users to unduly trust models \cite{lakkaraju2020fool, lipton2018mythos, ghorbani2019interpretation}, and more \cite{freiesleben2023dear}. This raises the question: Do the falsehoods and approximations (i.e. idealizations) operating in xAI have the same secret of success as the ideal gas law? Or are the critics largely correct that current xAI methods need improvement before systematic deployment, like the Dutch nitrogen case? When are deviations from the truth successful idealizations? When are they deceptive distortions? 

This paper seeks to animate a new research program for the study of xAI \textit{idealizations}. Idealization evaluation moves beyond concepts of falsehood or model fidelity capturing a deeper issue: the norms and practices surrounding model limitations and their context of appropriate use. I introduce a broad evaluation framework for idealization evaluation in xAI that aims to separate successful idealizations from deceptive explanations (SIDEs) (sect. \ref{SIDEs}). I identify areas where existing xAI evaluation methods are useful for idealization evaluation, and where innovation is necessary. To build this framework, I first take inspiration from the idealization practices in the natural sciences (sect. \ref{phil}). I take an off-the-shelf theory of idealization influential in philosophy of science, the minimalist view of idealization, and apply it to feature importance (sect. \ref{SIDEs}) and counterfactual explanation methods (sect. \ref{eval}), showing through a qualitative analysis that idealization failure is common. Lastly, I consider ways of ameliorating idealization failure and point to future research directions for building a theory of idealization for xAI (sect. \ref{success}). It is also important to say what this paper does \textit{not} do. It does not argue for a new philosophical theory of idealization or seek to diagnosis exactly what kind of idealization xAI methods engage in. Instead, I provide a modular framework for and show the potential for xAI researchers to engage with work on idealization in philosophy of science. The paper aims to make the following contributions: 

\begin{enumerate}
     \item Conceptualizes the field of xAI as solving an idealization problem
    \item Introduces the SIDEs framework for evaluating idealizations in xAI, derived from normative foundations of idealization in the natural sciences and philosophy of science.
    \item Provides a theoretical grounding for novel evaluation methods for xAI.
    \item Identifies possible novel practices of idealization in xAI that needs normative analysis. 
\end{enumerate}


\section{Background and Related Work} \label{related work}


The ever-growing fingerprint ML has on the production of scientific and social knowledge comes with challenges. One often cited issue is model opacity \cite{creel2020transparency, burrell2016machine, boge2022two}. Transparency of ML decisions is important for building trust \cite{jacovi2021formalizing, MILLER20191, lipton2018mythos, grote2021trustworthy}, it might be legally required \cite{selbst2018meaningful, goodman2017european}, and convincing arguments have been made for a moral right to explanation in high-impact contexts \cite{vredenburgh2022right}. This need for transparency inspired a proliferation of different interpretability and xAI techniques to solve the problem of opacity by providing insight into the reasons behind ML classifications. Post-hoc feature importance methods remain the most influential approach in xAI. These methods seek to approximate how much particular features contribute to the model's decision locally around each prediction. Examples include LIME \cite{ribeiro2016should}, SHAP which utilizes coalitions game theory \cite{lundberg2017unified}, and saliency maps that visualize regions of interest \cite{simonyan2014deep, ancona2017towards}. These methods differ from example-based or decision-rules \cite{molnar2020interpretable}, and differ from global explanation methods that seek to capture the behavior of a black-box system as a whole \cite{lakkaraju2019faithful, ibrahim2019global}. However, there are ways to use LIME and SHAP to get close to a global explanation by aggregating many local explanations. 
Counterfactual explanation (CE) methods have recently gained notoriety as the leading alternative to feature importance methods \cite{venkatasubramanian2020philosophical, wachter2018counterfactual, mc2018interpretable, hendricks2018generating, barocas2020hidden, mothilal2020explaining, ustun2019actionable, russell2019efficient}. CE methods seek to answer \textit{what-if-things-had-been-different} questions by probing the ML model to see what minimal changes would reverse the ML decision. 
There are a variety of different algorithms for generating or filtering which counterfactuals would be relevant in different contexts and for different stakeholders (see \cite{karimi2020survey} for a review). Despite these acheivements there remain conceptual and evaluative challenges to xAI and explainability.

\paragraph{\textbf{Conceptual issues for xAI}}
There are several conceptual contributions that philosophers of science have made in debates around xAI. Most notably, on the concept of \textit{explanation} \cite{MILLER20191}. Central questions concern what type of information is required to fulfil the definition of an `explanation' in philosophy of science  \cite{buijsman2022defining, erasmus2021interpretability, paez2019pragmatic, nyrup2022explanatory, MILLER20191}, such as conforming to a covering-law view of explanation \cite{erasmus2021interpretability} or having an additional link between the model and the world \cite{sullivan2022understanding}. Others focus on ethical considerations regarding whether certain xAI methods can fulfil moral requirements for explanation \cite{vredenburgh2022right, venkatasubramanian2020philosophical}, such as a principled reason explanation \cite{barocas2020hidden}.  These normative based approaches have exposed various challenges with xAI. For example, \citet{symons2022epistemic} discuss the issue of epistemic injustice in the context of trying to solve ML opacity. Others have suggested that CE methods have the potential to hide bias \cite{barocas2020hidden, sullivan2022explanation, aivodji2019fairwashing, sullivan2019ideal}. While even some argue that xAI methods are unnecessary and that model evaluation should focus instead on notions of reliability \cite{duran2021afraid, duran2018grounds, london2019artificial, grotereliability}. Discussions of  the norms of \textit{explanation} are no doubt important and necessary. However, as I hope to show in this paper, norms of explanation are distinct from the norms and ideals that govern idealization. However, only recently has their been a suggestion that the concept of idealization in xAI may be useful \cite{beisbart2022philosophy, fleisher2022understanding}. Nevertheless, here researchers stop short of discussing how xAI researches could actually use idealization theory, and how we could evaluate idealizations.


\paragraph{\textbf{Current approaches to xAI evaluation}}
Evaluating xAI methods in computer science include experimental methods, such as comparisons of accuracy and model fidelity between different algorithms and benchmarks \cite{BARREDOARRIETA202082, lipton2018mythos}, whether methods are robust under manipulation or perturbations \cite{slack2020fooling, slack2021counterfactual, dombrowski2019explanations, alvarez2018robustness}, or whether such methods conform to human expectations \cite{gilpin2018explaining, mittelstadt2019explaining, nauta2022anecdotal}. Current evaluation methods have exposed a number of vulnerabilities. Accuracy tests conducted on feature importance methods, found that the best performing method only approached 85\% agreement with the black-box model, with LIME often scoring lower \cite{lakkaraju2019faithful}. Furthermore, feature importance methods were found to be vulnerable to adversarial manipulation. \citet{slack2020fooling} were able to create explanations that hid the most salient feature for classification for SHAP and LIME. \citet{ghorbani2019interpretation} found such methods were highly sensitive to small changes to input data, with others finding that they are not able to capture causal notions \cite{lipton2018mythos, rudin2019stop, plumb2018model}. 
Counterfactual explanation methods can also fall prey to manipulation. Specifically, it was found that hill-climbing CE methods can converge to different local minima resulting in possible manipulation \citep{slack2021counterfactual}. They also suffer from the rashomon effect where different counterfactuals explaining the same decision can be contradictory \cite{carvalho2019machine, leo2001statistical}. 

While current methods of xAI evaluation are no doubt insightful, important gaps remain. Current experimental methods actively look for vulnerabilities and look for cases where methods break down or where new methods show an improvement compared to benchmarks. However, they stop short of providing an analysis that evaluates whether the \textit{limitations} and the distortions xAI methods introduce are actually problematic or could be a case of a successful idealization. Instead, several critics simply point to the existence of possible manipulation and limited model fidelity, as itself a strike against the method \cite{rudin2019stop, krishna2022disagreement, lipton2018mythos, lakkaraju2020fool, alvarez2018robustness}. While others have argued that current notions of fidelity are ill-equipped to capture cases of misleading explanation  \cite{lakkaraju2020fool, BARREDOARRIETA202082}. 
The potential for misleading explanations has also inspired a user-centered approach to evaluation, where xAI evaluation is geared toward fulfilling either actual user preferences, or expected user perceptions of usefulness \cite{jacovi2021formalizing, watson2021local, rong2022towards}, including the introduction of normative stakeholder sensitive frameworks {\cite{doshi2017towards, mohseni2021multidisciplinary, zednik2021solving, kasirzadeh2021reasons, langer2021we}. Again, while this evaluative approach is important, normative and theory-based evaluations concerning the potentially \textit{positive} role a lack of model fidelity could have for xAI methods are lacking.


\paragraph{\textbf{Closing Gaps}}

 This paper aims to address the above gaps by introducing the concept and framework of idealization evaluation. While model auditing techniques look for vulnerabilities, idealization evaluation asks whether such vulnerabilities are problematic or actually a successful tool. Moreover, idealization evaluation provides the conceptual tools for identifying the fundamental goals for xAI more so than just looking at theories of explanation.
With this paper, I hope to show the need for xAI research to build a theory of idealization and engage directly in idealization evaluation. Without a proper theory of idealization, it remains difficult to thoroughly diagnose the success of xAI methods. Idealization is inevitable, but if done right, idealization is desirable. My approach in this paper takes inspiration from idealization in the natural sciences and the philosophy of science to gain insight into how researchers can begin the project of idealization evaluation in xAI. I return to how idealization evaluation fits within current xAI research in section \ref{SIDEs}.

\section{Idealization in the Natural Sciences and Philosophy of Science} \label{phil}
Idealizations are the (intentional) distortion of real-world features that are present in a model or theory. In science idealizations are many. Examples include the ideal gas law and frictionless planes in physics, perfectly rational agents in economics, infinite populations and the absence of genetic drift in biology, etc. The way philosophers of science understand the concept of idealization might be best illustrated with an example outside of science. The Tube map of London’s underground has neatly organized lines, and the circle line resembles a circle. However, the Tube map distorts the actual layout of the Tube tunnels. In reality, the interconnection of tunnels is complex and rarely a straight line \cite{purtill15}. The official Tube map does more than leave out detail; it intentionally distorts the real layout of tunnel paths. The Tube map idealizes London's subway structure. Philosophers of science have sought to understand and conceptualize the nature, function, and epistemic value of idealizations in scientific inquiry \cite{elgin2017true, sullivan2019ideal, lawler2021scientific}.

\input{tabels/definitions.tex}
\paragraph{\textbf{Features and types of idealization practices:}}
In a landmark paper in philosophy of science, \citet{weisberg2007three} proposed that idealizations should be categorized by their specific scientific practices made up of the activity of scientists, the norms or values that govern these practices, and how these norms are justified. We can translate these aspects of idealization into three features (\textsc{idealization practice}, \textsc{ideals}, \textsc{rules}) that are important in the natural sciences, adding two additional features for xAI (See Table \ref{table:defs}).  Philosophers of science have conceptualized several different idealization practices in the natural sciences, like the unique quality of infinite idealizations \cite{shech2018infinite} and asymptotic idealizations in physics  \cite{batterman2001devil, strevens2019structure}, hypothetical-pattern idealization in biology \cite{rohwer2013hypothetical}, the practice of Galilean idealization, multiple-model idealization \cite{weisberg2007three}, and more. In this paper, we restrict discussion to one influential theory of idealization, Strevens' \cite{strevens2011depth, strevens2016idealizations} minimalist view of idealization (MinI). Below I will discuss how MinI works in a simple physics case, and in the next section discuss why MinI resembles the idealization practices in xAI and is a good place to start for evaluating xAI's idealizations. In section \ref{success}, I discuss alternative idealization practices that xAI could be engaged in.

\paragraph{\textbf{Minimalist idealization in physics}} The underlying norm for MinI is that simple models and explanations are better than more complex ones. Understanding phenomena requires isolating relevance from irrelevance often requiring idealization. MinI's \textsc{idealization practice} consists of devising scientific (or mathematical) methods for reducing the number of features that give rise to a phenomenon, highlighting the \textit{difference-makers}, and only distorting the \textit{non}-difference-makers. As such, the governing \textsc{ideals} for MinI are simplicity and isolating difference-makers \cite{strevens2011depth, strevens2016idealizations}. In physics, relevance and difference-making is usually a type of causal or dependency entailment, where some causal consequence can be (logically) derived from a set of initial conditions along with a causal law. In cases where the law is non-causal, the entailment is a different type of dependency entailment (such as a mathematical dependency) \cite{lange2016because}. While \citet{strevens2011depth} focuses on causal difference-making, others have adopted non-causal approaches to MinI \cite{batterman2014minimal}.

The most discussed example of MinI is the ideal gas law. The ideal gas law introduces the false assumption that a system consists of \textit{N} non-interacting particles so that physicists can clearly see that phase space is proportional to volume. The justification for adding the idealization of non-interacting particles is that in contexts of low pressure and high temperature, particle interactions are virtually insignificant to the relations between pressure, temperature, molarity, and volume. The idealization highlights this irrelevancy in a way that a more accurate representation hides. 

However, if we remove the idealization, we can still determine the irrelevance of particle interactions. We can derive the virial equation of state directly from statistical mechanics with arbitrary precision by extending the equation indefinitely, where each added term is derived from an increasingly detailed and accurate representation. However, the contribution of each added term becomes vanishingly small, again resulting in the ideal gas law \cite{sullivan2019ideal}. Thus, the ideal gas law satisfies the inclusion and fidelity \textsc{rules} of MinI by only removing and distorting aspects that do \textit{not} affect causal-entailment (i.e. only distorting non-difference-makers). 
In cases where entailment fails and the ideal gas law does not capture gas behavior (e.g. in high pressure), other laws are required (e.g. Van der Waals). Importantly, even if it is possible to de-idealize in this case, MinI is still appropriate. MinI captures the difference-makers that a more accurate alternative does not. As such, idealized distortions are permanent fixtures--even if they can in principle be removed--because they distinguish relevance from irrelevance.

\paragraph{\textbf{Idealization in xAI?}}  Machine learning is not the type of practice that philosophers of science have built their idealization theories around. ML models are complex instead of simple and they are not constructed with built-in theoretical assumptions where model equations explicitly represent processes in the target system \cite{knusel2020understanding}. ML models are often used precisely because such theoretical assumptions are unavailable, or because researchers are interested in prediction or overlooked patterns of interest. Moreover, in the case of xAI, the xAI model is an idealization not of the world but another model (the ML model). Thus, we need to separate between two questions of xAI and ML idealizations: 

\begin{itemize}
    \item \textbf{\textsc{Model-World} question}: How do black-box ML models idealize some real-world phenomenon? (e.g. how do ML models idealize aspects of disease indicators?) \cite{duede2022deep, sullivan2023machine, sullivan2022understanding}
    \\
    \item \textbf{\textsc{Model-Model} question:} How do xAI methods idealize how a black-box ML model works? How is an xAI method an idealized representation of the black-box model? (e.g. how do feature importance methods idealize aspects of the ML model decision process?)
\end{itemize}

XAI mainly concerns the \textsc{Model-Model} question (we return to \textsc{Model-World} questions for xAI in sect. \ref{eval}). Like the ideal gas law, there are several similarities between the xAI project and MinI. Current work in xAI often describes the ultimate goal of xAI methods as uncovering how black-box models make decisions, capturing how various inputs can cause a particular output in the black-box model \cite{MILLER20191, watson2022conceptual, plumb2018model}. This leaves \citet{fleisher2022understanding} to argue that feature importance methods are a kind of MinI because they satisfy simplification, flag difference-makers, and focus on a specific causal pattern in the black-box model that gives rise to the decision (i.e. answering the \textsc{model-model} question). For example, LIME uses linear approximation methods that distort aspects of the black-box model decision making process, but does so by aiming to find the features that are the central difference-makers for a given local decision (e.g. high debt is the largest difference-maker for why the black-box model recommended loan rejection). But are feature importance successfully engaging in MinI, like the ideal gas law?

\section{Toward a framework for idealization evaluation in xAI} \label{SIDEs}

In this section, I introduce the SIDEs framework. SIDEs consists of a high-level modular workflow (Figure \ref{fig:SIDEs}) that can guide researchers with key questions for reflection and qualitative evaluation of xAI idealizations. I go through each phase of SIDEs, identifying areas where existing theories and experimental evaluation techniques are useful and where innovation is required. If idealizations meet the standards for each phase, with alignment between phases, then the idealization is successful. Idealization failure occurs when there is misalignment or the idealization fails to meet the standards for a given phase. Throughout this section, leading feature importance methods, LIME and SHAP, are qualitatively evaluated to illustrate how SIDEs can identify risks of idealization failure. In section \ref{eval} I turn to CE methods.

\begin{figure}[ht]
\centering
\includegraphics[scale=0.60]{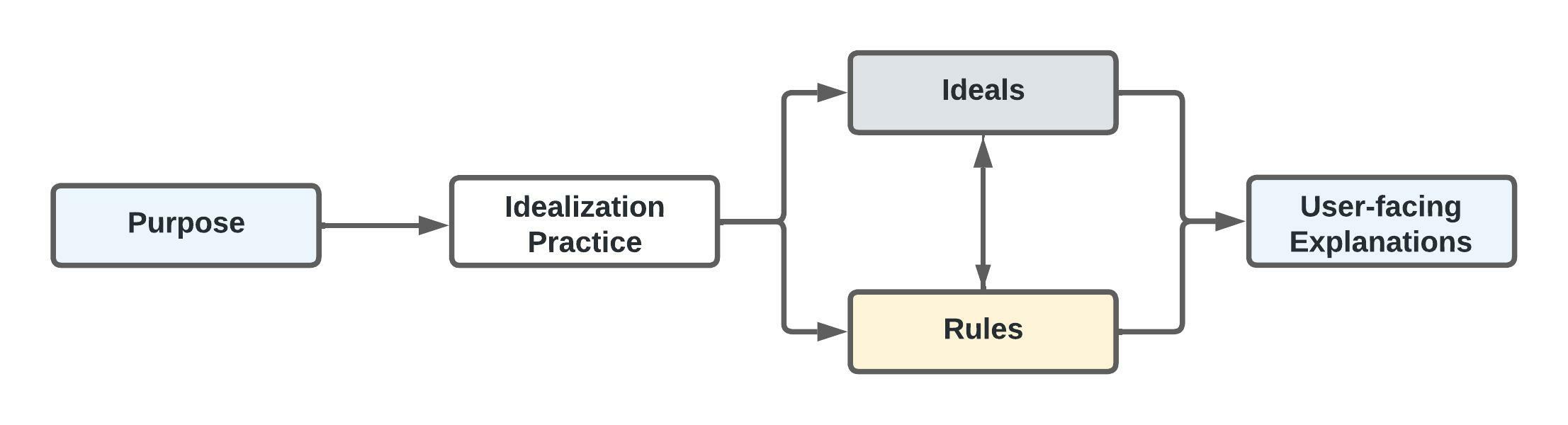}
\caption{SIDEs Workflow}
\Description[]{The SIDEs workflow from left to right consists of identifying purpose, idealization practice, ideals and rules, and ends in user-facing explanations}
\label{fig:SIDEs}
\end{figure}

\subsection{Purpose}
In philosophy of science idealization analysis begins with an idealization practice. However, xAI calls for starting idealization analysis with the purposes that researchers are aiming to achieve with idealizations \cite{buchholz2023means}. In the natural sciences, idealizations are discussed in the context of ideal scientific agents, so the central purpose of an idealization is presupposed to be epistemic by enabling understanding of phenomena \cite{rice2019models, potochnik2017idealization}. However, xAI methods serve a variety of purposes beyond epistemic purposes, including ethical purposes like recourse, where the aim is informing end-users about actionable changes that could reverse a negative decision \cite{ustun2019actionable}. To capture this difference, the SIDEs workflow starts with identifying the overall purpose the xAI method serves in a given context. 

I want to highlight two potentially conflicting purposes of xAI methods: epistemic and ethical purposes. However, additional purposes are possible (e.g. legal compliance). While it can be helpful to discuss purposes of xAI broadly, finer grained aims are more useful for idealization evaluation. For example, fine-grained epistemic purposes could be understanding the phenomena the model bears on \citep{sullivan2022understanding}, predictive knowledge \cite{integrating-prediction-and-explanation}, user-specific epistemic goals \citep{sullivan2019reading}, and more \cite{parker2020model, bokulich2021data}. Ethical purposes include fairness \cite{balagopalan2022road}, providing users with recourse (i.e. actionable interventions) \cite{venkatasubramanian2020philosophical},  exposing model bias, etc. Purposes are not necessarily mutually exclusive and may overlap. 
A single xAI explanation might aim to satisfy several purposes at once, such as providing users understanding of the model and building users' trust in the model, while also providing a user with algorithmic recourse. SIDEs does not preclude xAI methods from serving multiple purposes; however, the bar for successful idealization could become considerably higher (see section \ref{eval}). 

\subsubsection{Evaluating Purpose}
The \textsc{purpose} phase in SIDEs asks researchers to reflect on what purposes an xAI method does and \textit{should} have in a specific context. There are several existing works that have identified various purposes that xAI methods serve \cite{lipton2018mythos, krishnan2020against}, and there is normative and theory-based work on xAI concerning what purposes xAI should have that can be helpful for evaluating \textsc{purpose} \cite{kasirzadeh2021reasons}. A central pitfall for xAI methods is the risk of idealization failure due to misalignment with purpose. For example, in the original LIME paper, \citet{ribeiro2016should} describe the motivating purpose behind LIME as establishing user trust. However, trust could have divergent underlying aims. On the one hand, trust can serve an ethical purpose. In clinical cases, patients trust of a doctor's diagnosis is often not grounded in the patient's knowledge or understanding of the diagnosis \cite{millum2021informed, wolfensberger2019trust}. On the other hand, there are other contexts where trust is only achieved when users understand the reasoning behind decisions. This tension between different functions of trust complicates the picture of whether an xAI method engages in idealizations that could fulfil these various purposes. For example, \citet{lakkaraju2020fool} found LIME is able to manipulate user trust.

\subsubsection{Role and limits of current work}
Current research directions in xAI are well-placed to evaluate the purposes that xAI methods do and should have, with significant work already being done \cite{ nyrup2022explanatory, jacovi2021formalizing, lipton2018mythos, krishnan2020against, buchholz2023means}. Re-conceptualizing xAI as an idealization problem relies on this work as the first fundamental step toward establishing how model fidelity should be understood and which features of black-box models can be distorted (i.e. idealized). Idealization evaluation asks researchers when analyzing purposes of xAI to consider the extent to which model fidelity matters.

Key questions for the \textsc{purpose} phase are:
\begin{itemize}
    \item \textit{What purpose does an xAI explanation have in a particular context? What purpose should it have?}
    \item \textit{What aspects of the model need to be known for a particular purpose?}
    \item \textit{\textbf{Example}: What notion of trust is an explanation aiming for in a particular context?}
    \item \textit{\textbf{Success}: The purpose of the xAI explanation is appropriate for the given deployment context.}
\end{itemize}

\subsection{Idealization practices}
As discussed in section \ref{phil}, idealization practices consist of the set of scientific methods and practices along with the justification of those practices. One central research area in philosophy of science is conceptualizing different idealization practices across the sciences into distinguishable types or theories of idealization. 

\subsubsection{Evaluating Idealization Practices}
Evaluating the \textsc{idealization practices} phase involves two central aspects. The first is a descriptive project that systematizes current work in xAI, elucidating a set of common aims and methodologies. This can be done for xAI in general or for a specific class of xAI methods. Second, evaluating \textsc{idealization practices} consists of a justification step that can ground the legitimacy of the idealization practice. Using MinI as our working hypothesis, the methodology of MinI consists in omitting or distorting (causal) influences for the purposes of highlighting the central (causal) difference-makers or (causal) patterns. MinI is justified both through a strong conceptual foundations in scientific understanding, explanation, and (casual) difference-making, and in its empirical success \cite{lawler2021scientific}.

\subsubsection{Role and limits of current work}
Currently there has been very little work trying to conceptualize the type of idealization practices computer scientists are engaged in when developing xAI methods \cite{fleisher2022understanding}, and these practices are still arguably elusive \cite{lipton2018mythos}. However, as discussed in sect. \ref{phil}, there are several similarities between current xAI methods and MinI. XAI aims to cut through the noise of many feature interactions to arrive at the chief difference-makers for a decision. For the purposes of this paper, we evaluate xAI methods as if they are engaging in the idealization practice of MinI. However, this paper calls for xAI to actively engage in solidifying one or more idealization practices and to work with philosophers of science on establishing conceptual foundations that justify these idealization practices. This is a central area that requires innovation and future research (see also section \ref{success}).

Key questions for the \textsc{idealization practices} phase are:
\begin{itemize}
    \item \textit{What are the specific methods of deriving idealizations (e.g. introducing certain idealization assumptions, mathematical operations applied on data that results in  distortions of the phenomena)?}
    \item \textit{Does the idealization practice align with the purpose of the idealized model?}
    \item \textit{What justifies this particular idealization practice? Why is  it suitable for the identified purpose?}
    \item \textit{\textbf{Example:} Minimalist idealization provides better understanding of the relevant difference-makers.}
    \item \textit{\textbf{Success:} The idealization practice is well-grounded and justified in a specific domain, aligning with \textsc{purpose}.}
\end{itemize}

\subsection{Ideals and Rules}
\textsc{Ideals} are the norms and values that govern a specific idealization practice. \textsc{Rules} are the operationalization of these norms and values. For example, the ideals for MinI are to isolate the minimum number of difference-makers that capture a phenomenon, while Strevens \cite{strevens2011depth} describes the fidelity and inclusion rules of MinI as satisfying a `causal entailment' test, where modelers remove (or distort) features, finding the minimal amount that still entail the desired event.

\subsubsection{Evaluating \textsc{ideals} and \textsc{rules}}
Evaluating idealizations in the \textsc{ideals} and \textsc{rules} phase requires validating whether specific rules embody target ideals, and analyzing trade-offs between different ways of operationalizing ideals. Idealizations can be experimentally validated by developing tests that satisfy these rules. Adopting our working hypothesis that xAI is engaging in MinI means that for xAI methods to be a legitimate idealization of a black-box model and pass the \textsc{ideals} and \textsc{rules} phase, xAI methods must isolate the difference-makers for the target black-box model's decision by undergoing a (causal) entailment test that ensures the xAI method gives the same results as the target black-box model. Passing an entailment test would mean, in theory, that the xAI method uncovers the minimum set of difference-makers that determine the black-box model's decision \cite{watson2022conceptual}, and thus any distortion it makes in the process is a legitimate one.

However, specifying an adequate (causal) entailment test is not simple. First, the notion of difference-making in xAI must be clear. For \citet{strevens2011depth, strevens2016idealizations}, MinI aims for causal difference-making, where the notion of causality is left implicit and entailment more closely resembles logical deduction. Comparatively, philosophical theories like an interventionist framework \cite{woodward2005making, watson2022conceptual} or a counterfactual framework \cite{lewis1973counterfactuals, kasirzadeh2021use} would result in grounding different causal rules. Alternatively, MinI need not aim for \textit{causal} difference-making at all; other notions of difference-making are consistent with MinI (logical, probabilistic, mathematical, etc.). Thus, it is necessary to establish the specific ideals that a given idealization seeks to capture. Second, even once we settle on a notion of difference-making for MinI, there are still different possible rules that could capture the norms of MinI and serve as a basis for idealization evaluation. In this section, I consider three possible rules based on Strevens view of MinI. My aims are to 1) illustrate how ideals might be operationalized (see Table \ref{table:rules}); 2) discuss how to think about trade-offs and whether a certain rule embodies the target ideal; and 3) discuss where existing evaluation methods in xAI are useful and areas where innovation is needed.  

\input{tabels/causal_rule}

First, MinI entailment could be a global entailment where the xAI model shares the same mapping of model inputs to outputs as the black-box model. The \textsc{map} rule requires that the mapping of inputs and outputs for an xAI model, \textit{e}, is the same as for the black-box model, \textit{b}. \textsc{Map} does not look at the features that the xAI model highlighted as relevant. As long as there is a 1 to 1 input-output mapping, then this is enough to establish a global notion of (causal) entailment. \textsc{Map} aligns with some of the current approaches to xAI evaluation. Accuracy and model fidelity metrics used for feature importance methods aim to see how well explanations mirror black-box predictions \cite{lakkaraju2020fool}. The results of these tests have exposed important vulnerabilities. For example, \citet{lakkaraju2019faithful} found the best performing method only approached 85\% agreement with the black-box model, with LIME often scoring lower. 
Even if we have a 1-1 mapping, an important trade-off to consider with \textsc{map} is how well it aligns with the purpose for xAI. Usually the purpose of an xAI method is for users to learn about the reasons for why the black-box model made its decision, not merely that an alternative proxy-model (in the case of LIME and SHAP) can derive the same predictions, which is why \citet{lakkaraju2019faithful} argue that high-fidelity xAI models aren't enough.

Second, as an alternative, we could operationalize entailment more closely with Strevens' \cite{strevens2011depth} own elimination test for MinI in the natural sciences. \textsc{elimination} tells us that for all the features \textit{Y} that are in the set of putatively irrelevant features \textit{I} found from the xAI model \textit{e}, we can remove those features from the set of all input features \textit{X} from the black-box model \textit{b} and still receive the same decision. According to \textsc{elimination}, evaluation could occur for any given local decision to see whether in that instance there is entailment between the black-box model and the xAI model. As we saw, one general downside of \textsc{map} is that an input-output pairing does not capture which features xAI methods determine are (ir)relevant \textsc{elimination} captures this aspect of explainability. Since \textsc{elimination} can be evaluated per local decision, there is more flexibility for success. Some local decisions may not satisfy \textsc{elimination}, while others do. Indeed, in the cases where the xAI method works well, \textsc{elimination} should be satisfied. But the trade-off here is that idealization failure would needs to be tested for each local decision. 

Lastly, as yet another alternative, some philosophers of science have argued only a probabilistic notion of (causal) relevance or difference-making is necessary for MinI \cite{potochnik2015causal}. \textsc{Prob} can apply to any other rule. It says that the probability that a rule applies to x is greater than some probability threshold \textit{t}. That said, \textsc{prob} may not align well with the ideals for MinI for many xAI purposes, for example users may not want to know what the most probable reason for the decision was, but the actual reason for the decision. However, when using xAI for the purposes of de-bugging or de-biasing a black-box model, \textsc{prob} could be appropriate.


\subsubsection{Role and Limits of Current Work}
The \textsc{ideals} and \textsc{rules} phase is the area in xAI idealization evaluation that requires innovation. In philosophy, \citet{fleisher2022understanding} argues feature importance methods are a kind of MinI, but he argues this on the level of \textsc{ideals} and stops short of discussing whether particular xAI methods actually succeed at MinI instead of merely \textit{aiming} for MinI. Other works focus on the norms and ideals of \textit{explanations} for xAI. Citing one example, \citet{watson2021local} propose the ideal of sufficiency for xAI methods because they provide potentially more useful and `lower cost' explanations for users. However, the norms and ideals for explanation are distinct from the norms and ideals that govern idealization evaluation. Idealization evaluation, first and foremost, treats xAI models not as explanation tools, but as idealized models of more complex models. SIDEs evaluates whether the distortions a given xAI method engages in succeeds at living up to the purpose and norms of idealization theory.

On the level of \textsc{rules}, there are existing experimental techniques that have been used to evaluate LIME and SHAP that capture some of the spirit of our suggested entailment rules for MinI, like \textsc{map} discussed above. While LIME and SHAP do not satisfy \textsc{map} because the accuracy rates do not reach perfect alignment between the black-box model and the xAI model, \textsc{map}'s \textsc{prob} counterpart has some level of success depending on the probability threshold. However, even a .8 probability may be too low to establish a strong sense of causal entailment for MinI. So if the ideal for xAI is a causal ideal then there is still idealization failure in the leading feature importance methods. There is no experimental test that captures \textsc{elimination} that I am aware of. However, feature importance methods are vulnerable to adversarial manipulation. \citet{slack2020fooling} were able to create explanations that hid the most salient feature for classification for SHAP and LIME. \citet{ghorbani2019interpretation} found such methods were highly sensitive to small changes input data. Others have found that they are not able to capture causal notions \cite{lipton2018mythos, rudin2019stop, plumb2018model}. 
This type of manipulation suggests that feature importance methods are not robustly conforming to \textsc{elimination}. Moreover, adversarial examples where the adversarial classifier achieves strong \textsc{prob(map)} but the perturbed instances are different (see \cite{slack2020fooling}) shows the tension between more global oriented rules like \textsc{map} and local entailment rules like \textsc{elimination}. This suggests that in order to fully capture MinI, it may be necessary to satisfy both \textsc{map} and \textsc{elimination}.

Innovation on new experimental evaluations that instantiate idealization rules could be promising. In this paper, I took just one conception of (causal) entailment from Strevens to ground potentially new experimental evaluation tests for xAI.  XAI researchers should consider the idealization norms and ideals they are aiming to achieve and align experimental evaluation tests with these norms. Existing theories of idealization in philosophy of science could serve as inspiration for establishing quantitative evaluation metrics for idealization.

Key questions for the \textsc{ideals} and \textsc{rule} phase are:
\begin{itemize}
    \item \textit{What are the norms and values that govern and justify an idealization practice?}
    \item \textit{What are the possible ways to operationalize ideals to experimentally and formally evaluate whether an idealized model satisfies the ideals of the idealization practice?}
    \item \textit{What are the trade offs between different rules?}
    \item \textit{\textbf{Example}: Different possible entailment rules for MinI and the limitations of each.}
    \item \textit{\textbf{Success}: \textsc{Rules} adequately reflect \textsc{ideals}. The xAI method satisfies the rule by passing an experimental test.}
\end{itemize}

\subsection{User-facing explanations} 
Work on idealization in the natural sciences considers scientists as stakeholders. The ideal gas law idealization is successful mainly because the intended audience generally knows enough about physics to understand where the idealizations lie. XAI, on the other hand, serves many diverse stakeholders, many of which do not know how ML works in any detail. Therefore, attention must be paid to the way idealizations are presented to different stakeholders through \textsc{user-facing} explanations. Thus, the last step for the SIDEs framework is evaluating user-facing explanations.

}
\subsubsection{Evaluating \textsc{user-facing} explanations}
Evaluating the explanations that target users receive involves user testing and user studies to ensure that explanations align with \textsc{purpose} and do not mislead users with its idealizations. One general pitfall for aligning user-facing explanations with user values is the potential for `explanation hacking' where xAI methods are so flexible to display only those that are agreeable to users \cite{sullivan2024hacking}, which can result in fairwashing \cite{aivodji2019fairwashing}. Even if the xAI method is a successful idealization from a scientific or mathematical perspective (i.e. passes the \textsc{rule} phase), there can be idealization failure if it misleads users through explanation hacking or fairwashing techniques. Moreover, \citet{mittelstadt2019explaining} found that users found LIME and SHAP unintuitive. Other considerations include doing user-study research not just for experts, but for a variety of users that have different background assumptions \cite{ehsan2021expanding}. Including users from the global south, which are often ignored \cite{okolo2022making, peters2024weird}.

\subsubsection{Role and Limits of Current Work} Several existing works address issues regarding user-facing xAI. Some provide frameworks that incorporate stakeholder interests \cite{doshi2017towards, mohseni2021multidisciplinary, zednik2021solving, kasirzadeh2021reasons, langer2021we}, others explore how different explanation types affect user-trust \cite{jacovi2021formalizing}, cognitive bias \cite{bertrand2022cognitive, cau2023effects}, and understanding \cite{10.1145/3397481.3450644, mothilal2020explaining}. SIDEs adds to this by highlighting the need for user-facing explanations to align with the purpose of \textit{idealizations}, conveying the ideals and norms of the idealizations in the explanations they receive. For example \citet{schneider2023explaining}, used language similar to MinI when describing the results of their user study, saying that users could tease out what was relevant and irrelevant to a model decision.

Key questions for the \textsc{user-facing explanation} phase are:
\begin{itemize}
    \item \textit{Do users understand the purpose of the explanation?}
    \item \textit{How can user-facing explanations convey that the explanation is an idealization?}
    \item \textit{Are user-facing explanations evaluated in terms of the purpose? Are they evaluated in terms of another purpose?} 
    \item \textit{\textbf{Example}: A user study asks users about their impressions regarding the purpose and limits to the explanation.}
    \item \textit{\textbf{Success}: User-facing explanations align with \textsc{purpose} and users acknowledge the limited scope of explanations.}
\end{itemize}

\section{Evaluating Counterfactual Explanation Methods} \label{eval}
In the previous section, I introduced the SIDEs framework and presented a qualitative evaluation of leading feature importance methods, LIME and SHAP. In this section, I identify risks of idealization failure in counterfactual explanation (CE) methods, using SIDEs. It is important to note that proponents of CE methods boast that CE cannot be false since counterfactuals are generated from the black-box model itself \cite{mothilal2020explaining}. However, such methods still engage in idealization. CE methods must select which counterfactual scenarios are the most salient from a larger set of possible counterfactuals and implicitly rely on notions of difference-making that seek to tease out the relevant counterfactual scenarios from the less relevant. This is one reason why it is important to re-conceptualize xAI as seeking to solve an idealization problem, instead of the current frame of xAI aiming at `faithful' explanations.

\paragraph{\textbf{Misalignment in CE:}}
\citet{wachter2018counterfactual} highlight three different purposes for counterfactual explanation methods: \textit{i)} explain why a decision was reached, \textit{ii)} provide grounds to contest the decision, and \textit{iii)} provide users with actionable changes to reverse the decision. SIDEs first requires that the idealizations used align with each specific purpose. \textit{i)} has a clear epistemic purpose aimed at gaining understanding of the black-box model behavior, while using CE for \textit{iii)}--known as recourse--has gained considerable attention regarding its ethical promise \cite{venkatasubramanian2020philosophical, sullivan2022explanation}. A recourse explanation is one where users are given feasible actions for them to undertake to reverse a model decision (e.g. paying down existing debt to qualify for a loan). Importantly, recourse and purely epistemic explanatory aims come apart \cite{konig2023improvement, sullivan2024hacking}. Since recourse provides users with actionable changes that they can make to reverse a decision, recourse explanations can mask bias and principle-reason explanations \cite{barocas2020hidden}. It could be that an immutable feature, like gender or race, was \textit{the} biggest difference-maker for why the model made its decision. Such an explanation cannot, in principle, be a recourse explanation. This is one reason why \citet{sullivan2022explanation} suggest conceptualizing recourse as a recommendation to avoid misleading users.

The type of idealizations that can satisfy MinI for \textit{understanding the decision} do not immediately translate to idealizations that are acceptable if the purpose of the idealization is to provide users with recourse. For example, for the epistemic purpose of understanding, CE methods can idealize and distort the underlying causal structure in the data and idealize away any interdependence between features--especially if the underlying black-box model ignores interdependence between features. In the epistemic case, satisfying MinI only requires alignment between the xAI model and the black-box model (i.e. answering the \textsc{model-model} question). However, recourse explanations have a different target: the relationship between model features \textit{and the world} \cite{karimi2020survey}, thereby aiming to answer a \textsc{model-world} question. For recourse, an idealization that ignores feature interdependence and the underlying causal structure in data will likely fail MinI \textsc{rules} that are calibrated to capture aspects of real-world causal efficacy. Indeed, works have criticised CE methods that ignore feature interdependence as creating unrealistic advice \cite{hooker2021unrestricted}. 

\begin{figure}[ht]
\centering
\includegraphics[scale=0.46]{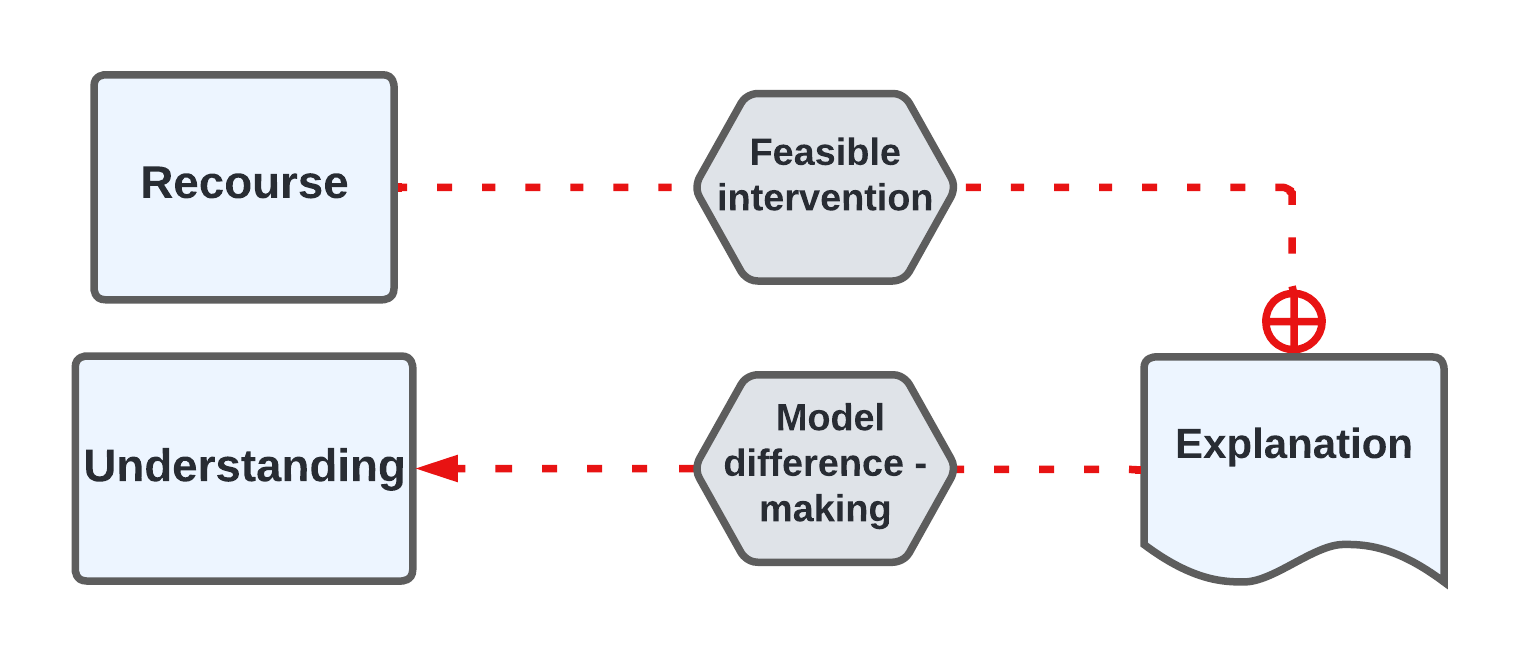}
\caption{Recourse Alignment Failure}
\Description[]{The purpose of recourse requires providing users with possible interventions on the model, but the user-facing explanation is presented to the user as providing information about model difference-making thereby fulfilling the purpose of understanding not recourse.}
\label{fig:recourse_mis}
\end{figure}


\textsc{Purpose}-alignment-failure can carry over to how xAI explanations are conveyed to users in the \textsc{user-facing explanation} phase. For example, recourse explanations are often presented as answering a \textsc{model-model} question (i.e. understanding of how the black-box model behaves), with several works evaluating their recourse method on whether users have similar understanding of the models decision boundary compared to, e.g. LIME \cite{mothilal2020explaining, wachter2018counterfactual, ustun2019actionable}. However, since the underlying purposes of recourse CE are feasibility and actionability, evaluating recourse CE should be done in terms of whether users find the recourse CE feasible. Understanding the model's decision is secondary. Figure \ref{fig:recourse_mis} shows this type of recourse CE alignment failure, where the explanation is evaluated based on the wrong \textsc{purpose.} This is not to say that recourse explanations could not be successful idealizations. SIDEs maintains that omitting or distorting the central reasons for a model's decision from a recourse explanation is legitimate so long as the \textsc{user-facing explanation} is aligned with actionability, while making clear to the user it is not an \textit{epistemic} explanation. Of course, it is possible that CE could satisfy both an epistemic purpose and recourse. However, for an idealization to achieve both aims there is a considerably higher bar where the CE method would need to satisfy both difference-making w.r.t. the \textsc{model-model} question and difference-making w.r.t. the \textsc{model-world} question. We should expect that this might be possible in some cases, but not likely in cases where immutable features are the largest difference-maker for a model's decision.

\paragraph{\textbf{\textsc{Ideal} and \textsc{rule} failure}}
Many of the same issues that come up with \textsc{rule} evaluation for feature importance methods also appear with CE methods. However, like the alignment issues discussed above, recourse CE has unique risks. The governing \textsc{ideals} for recourse are feasibility and actionability. Thus, a structural causal model (SCM) that captures how features within a model causally dependent on each other \cite{pearl2009causality, kusner2017counterfactual, baron2023explainable} will be necessary to satisfy MinI, albeit an idealized SCM. However, SCMs are far from attainable, requiring a link to the causal realities outside of the model \cite{sullivan2022understanding, baron2023explainable}. Thus, the \textsc{purpose} of recourse, coupled with the \textsc{idealization practice} MinI, requires a causal \textsc{rule} that can uphold the \textsc{ideal} of a SCM. However, \citet{karimi2020survey} find most works on recourse do not even aim for a SCM. Thus, these methods are engaging in idealization failure. \citet{karimi2020algorithmic}, on the other hand, employ a probabilistic approach to try and capture the ideal of a SCM with imperfect causal knowledge, and thus is a candidate for a \textsc{prob(scm)} rule, and might indeed be an idealization success, depending on how the \textsc{prob(scm)} rule isolates difference-makers. 

All told, researchers need to be mindful that \textsc{model-world} questions and \textsc{model-model} questions require different idealizations and have different idealization standards.

\section{Turning idealization failure into success} \label{success}

In our limited evaluation of feature importance and CE methods we found that on the working hypothesis that xAI methods are aiming for MinI there is widespread idealization failure, due to misalignment with \textsc{purpose} and \textsc{rule} failure, suggesting that these methods may likely distort more than just the non-difference-makers. Where does this leave us? In this section I discuss possible remedies for idealization failure and areas for future research.

\begin{figure}[ht]
\centering
\includegraphics[scale=0.46]{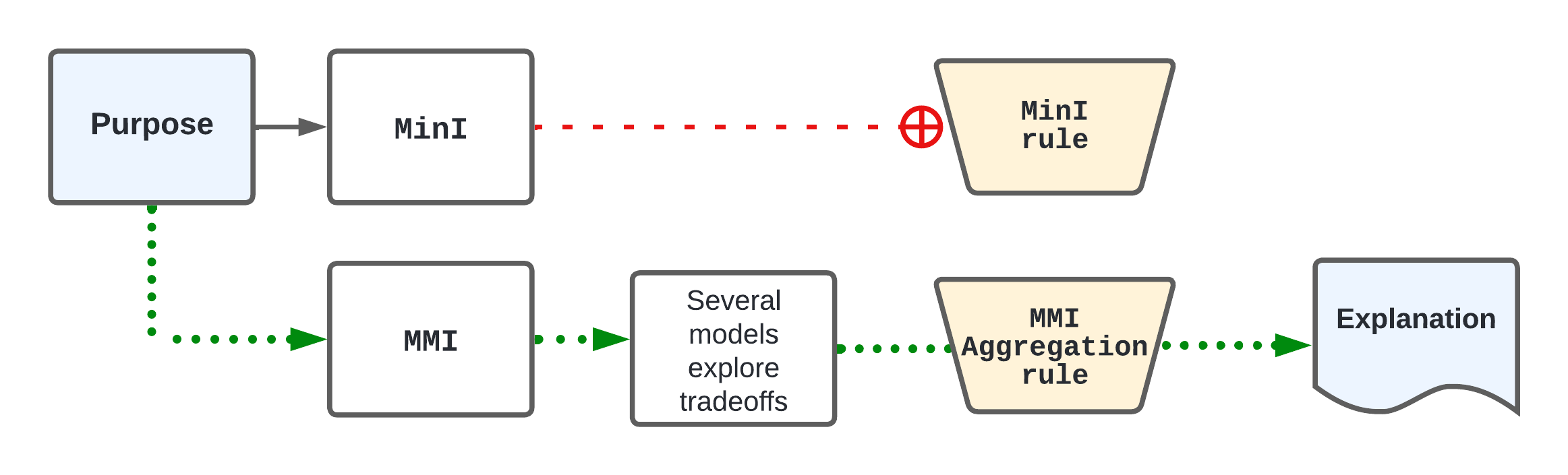}
\caption{Adopting an alternative \textsc{idealization practice} to address \textsc{rule} failure}
\Description[]{Minimalist idealization runs into rule failure. Re-aligning to multiple model idealization has the potential to become a success for explanation.}
\label{fig:mmi}
\end{figure}

\paragraph{\textbf{Adopting a different idealization practice:}}
If it is too difficult to satisfy one idealization practice, such as MinI, then idealization success can occur by creating alignment with a different idealization practice. \textsc{Rule} failure under one idealization practice does not entail \textsc{rule} failure under another. Figure \ref{fig:mmi} shows realignment with a different idealization practice after \textsc{rule} failure. As we said at the outset there are a number of idealization practices that philosophers of science have discussed. One alternative idealization practice that may be well suited for xAI is multiple-model idealization (MMI) \cite{weisberg2007three}. For some phenomena there may be several trade-offs that makes it either impractical or epistemically lacking to rely on just one model. MMI involves multiple models, with each model capturing one aspect or trade-off for some phenomenon. For example, chemists use both valence bond and molecular orbital models despite their incompatible assumptions \cite{weisberg2007three}. The underlying justification for MMI relies on the impossibility (either practically or necessarily) of a single model capturing maximal goals of representation, such as accuracy and generality \cite{levins1966strategy}. Furthermore, through robustness techniques among multiple models, MMI teases out relevance from irrelevance, something that MinI might fail to achieve in a single model \cite{Wimsatt2012-WIMRRA-2}. While each individual model in the MMI might have drawbacks, the collection of models taken together is able to provide understanding.

One notable solution to the idealization failure with MinI is to use LIME, SHAP and CE as part of a MMI instead. Realigning to MMI means that the \textsc{rules} for each xAI method can be less demanding. Each model, fulfiling a different \textsc{ideal} or \textsc{rule} explores different trade-offs, together capturing how a black-box model makes decisions. 
Moreover, MMI can be a temporary solution to xAI manipulation. \citet{slack2020fooling} found that adversarial attacks designed for LIME were ineffective against SHAP. Even though they also found that adversarial attacks designed against SHAP affected LIME, if the MMI also includes CE methods (or other methods) the vulnerability that each will be affected by the same attack diminishes. Furthermore, MMI can help with the problem of multi-purposes for xAI models. Users can be provided multiple explanations to fulfil these multiple purposes. However, MMI is not an simple fix. First, it is important to develop an aggregation \textsc{rule} for MMI idealizations. How should we weigh the different and sometimes conflicting models? This is a considerable undertaking. Second, current work has discussed that providing users with multiple different explanations can be counter-productive, creating confusion and cognitive overload \cite{poursabzi2021manipulating}. Thus, MMI might not be a useful idealization practice outside of an engineering model-auditing setting.


\paragraph{{\textbf{xAI, a novel idealization practice?}}}
In this paper, we looked at mature theories of idealization from the philosophy of science that were developed with the natural sciences in mind. It would not be surprising if those theories are altogether ill-suited for xAI since xAI and ML research has very different and specific requirements compared to the natural sciences. 
For example, one unique aspect of many xAI methods are their \textit{hyper locality}. In the natural sciences, idealizations often move away from local particulars to more global generalities. But current methods of xAI are distorting the global generalities of the black-box model to zoom in on a particular local explanation. Perhaps this is a novel \textit{hasty generalization} idealization practice. One area for future research is developing what this potentially novel idealization practice consists of and how it might be justified and grounded as a legitimate idealization practice. Importantly, the SIDEs framework cautions against idealization success simply by fiat (i.e. by claiming a new \textsc{idealization practice}). SIDEs requires a justification step for motivating why such a practice is legitimate.


\section{Conclusion}
XAI methods have received their fair share of criticism. However, with this paper I argued that one type of criticism--that xAI methods produce false explanations of black-box models--deserves closer attention. Specifically, this paper seeks to animate a new interdisciplinary research program in xAI that develops a theory of xAI idealizations and idealization evaluation. It is not simply departure from the truth that is problematic, but idealization failure. I introduced the SIDEs framework as a way for researchers to separate successful idealizations from deceptive explanations. SIDEs is a generalizable and modular conceptual framework that can guide researchers with key questions for reflection and qualitative evaluation, as well as provide the normative foundation for developing more concrete evaluative tests and benchmarks for xAI methods. SIDEs is primarily aimed at xAI researchers when developing and evaluating their methods (esp. the \textsc{ideals and rules} phase). However, there is also a place for SIDEs in more downstream uses. Those who deploy xAI models could use SIDEs in selecting which xAI method would be more successful for their purpose, such as providing recourse explanations vs. model auditing. However, this would require clear guidance on the results of the rest of the SIDEs workflow from xAI researchers. Moreover, the \textsc{user-facing explanation} phase could be useful for thinking about how to comply with the right to explanation in AI governance.

Using SIDEs, we found considerable risks of idealization with leading xAI methods. There are many ways in which current work in xAI is useful for idealization evaluation, such as identifying the purpose of xAI methods, developing user-centered studies to evaluate the efficacy of explanations. However, there are central ways in which innovation is necessary: 1) identifying and solidifying an idealization practice for xAI, including a justification for that practice, and 2) developing experimental tests that aim at evaluating how well an xAI method idealizes the target black-box model.






\section*{Acknowledgements}
 For helpful comments and discussions I'd like to thank Céline Budding, Thomas Grote, Yeji Streppel, Philippe Verreault-Julien, Carlos Zednik, with special thanks to John Mumm. I presented this work in various stages in development at the Munich Center for Mathematical Philosophy, University of Tübingen, Utrecht University, and Nicolaus Copernicus University. I thank those in attendance for the fruitful discussions that followed. This work is supported by the Netherlands Organization for Scientific Research (NWO grant number VI.Veni.201F.051).


\bibliographystyle{unsrtnat}
\bibliography{bib}

\end{document}

%% file: tabels/definitions.tex
\begin{table}[ht]
\small
\centering
\caption{Features of Idealization}
\label{table:defs}
\begin{tabular}{p{4cm}p{9cm} }
\toprule
\bf{Idealization 
Features}  & \bf{Description} \\
\midrule

\textsc{Purpose} & The purpose / function of the idealization (e.g. epistemic purpose, like understanding; ethical purpose, like recourse and contestability, etc.) 
\\\\ 
\textsc{Idealization Practice} & The set of scientific methods and practices that categorize a type of idealization, along with the justification of those practices  (e.g. Minimalist idealization) 
\\\\
\textsc{Ideals} & Values and norms underlying an idealization practice and that govern rule development (e.g. causal-entailment, feasibility for users)
\\\\
\textsc{Rules} & The way ideals are operationalized into a metric of evaluation
\\\\
\textsc{User-facing}
\textsc{explanations} & How idealizations are presented as explanations to end-users
\\ 
\bottomrule
\end{tabular}
\end{table}

%% file: tabels/causal_rule.tex
\begin{table}[ht]
\small
\centering
\caption{Entailment \textsc{rules}}
\label{table:rules}
\begin{tabular}{c @{\qquad\qquad} c @{\qquad\qquad} c}
\toprule

\textsc{map}  &  \textsc{elimination} &  \textsc{prob} \\\midrule

 $\forall x\; b(x)=e(x)$ &  $\forall Y \subset I_{e}\quad b(X-Y)=b(X)$ &   $\forall x\;  P(\operatorname{\textsc{rule}}(x)) > t$\\



\bottomrule
\end{tabular}
\end{table}